\documentclass[conference, 10pt]{IEEEtran}
\usepackage{amssymb,amsmath}
\usepackage{setspace}
\usepackage{graphicx}
\usepackage{epsfig,cite,latexsym,subfigure}
\usepackage{theorem}
\usepackage{times}
\usepackage{epsf,cite,subfigure}
\usepackage{amsmath,amssymb}

\usepackage{floatflt}
\usepackage{url}
\usepackage{times}
\usepackage[usenames]{color}
\usepackage{dsfont}
\usepackage{mathptmx}
\usepackage[final]{pdfpages}
\newcommand{\set}[1]{\left[#1\right]}
\newcommand     {\paren}[1]{\left(#1\right)}

\title{\vspace{-2mm}{\Large Adversarial Examples in RF Deep Learning:\\ Detection of the Attack and its Physical Robustness}}\vspace{-2mm}
\author{
\IEEEauthorblockN{\normalsize Silvija Kokalj-Filipovic, Rob Miller} \\
\IEEEauthorblockA{\small Perspecta Labs, Inc \\
\small\em \{skfilipovic, rmiller\}@perspectalabs.com}}
\begin{document}
\maketitle
\begin{abstract}
While research on adversarial examples in machine learning for images has been prolific, similar attacks on deep learning (DL) for radio frequency (RF) signals and their mitigation strategies are scarcely addressed in the published work, with only one recent publication in the RF domain \cite{Larsson}.  RF \emph{adversarial examples (AdExs)} can cause drastic, targeted misclassification results mostly in spectrum sensing/ survey applications (e.g. BPSK mistaken for OFDM) with minimal waveform perturbation.  It is not clear if the RF AdExs maintain their effects in the physical world, i.e., when AdExs are delivered over-the-air (OTA). Our research on deep learning AdExs and proposed defense mechanisms are RF-centric, and incorporate physical-world, OTA effects.  We here present defense mechanisms based on statistical tests. One test to detect AdExs utilizes Peak-to-Average-Power-Ratio (PAPR) of the DL data points delivered OTA, while another statistical test uses the Softmax outputs of the DL classfier, which corresponds to the probabilities the classifier assigns to each of the trained classes. The former test leverages the RF nature of the data, and the latter is universally applicable to AdExs regardless of their origin. Both solutions are shown as viable mitigation methods to subvert adversarial attacks against communications and radar sensing systems.
\end{abstract}
\section{Intro}
\begin{figure}[t] 
\vspace{-1mm}
\begin{center}
\hspace{-5mm} \includegraphics [width=3.5in]{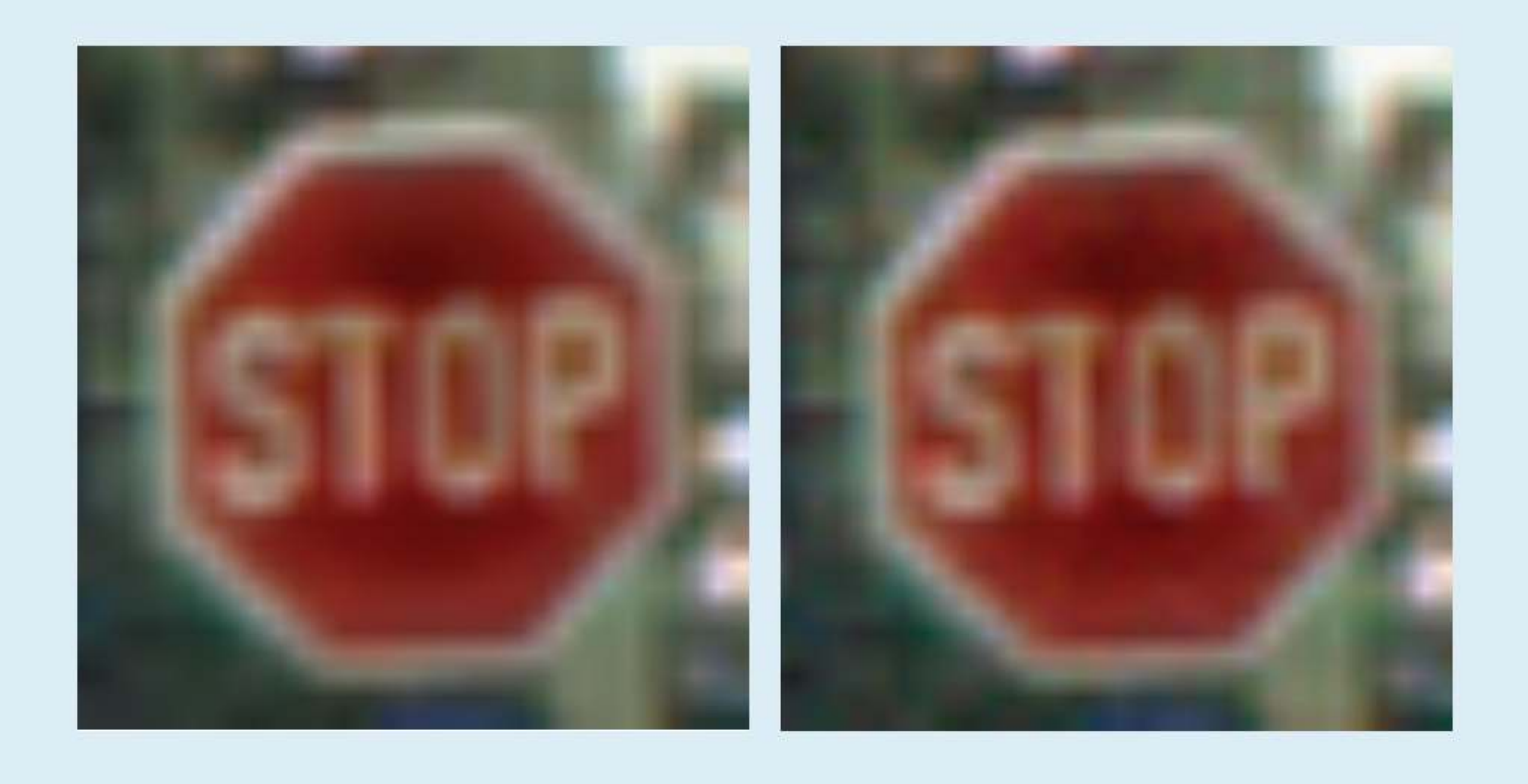} 
\caption{Stop-sign/ Yield-sign illustration of an adversarial image example against a DL classifier (e.g., implementing the vision at a driverless car): Here, the Stop Sign on the right is mis-interpreted as a Yield Sign}\vspace{-4mm}   \label{fig:f00}
\end{center}
\end{figure}

\begin{figure}[t] 
\vspace{-1mm}
\begin{center}
\begin{tabular}{c c}
\hspace{-10mm} \includegraphics [width=1.5in]{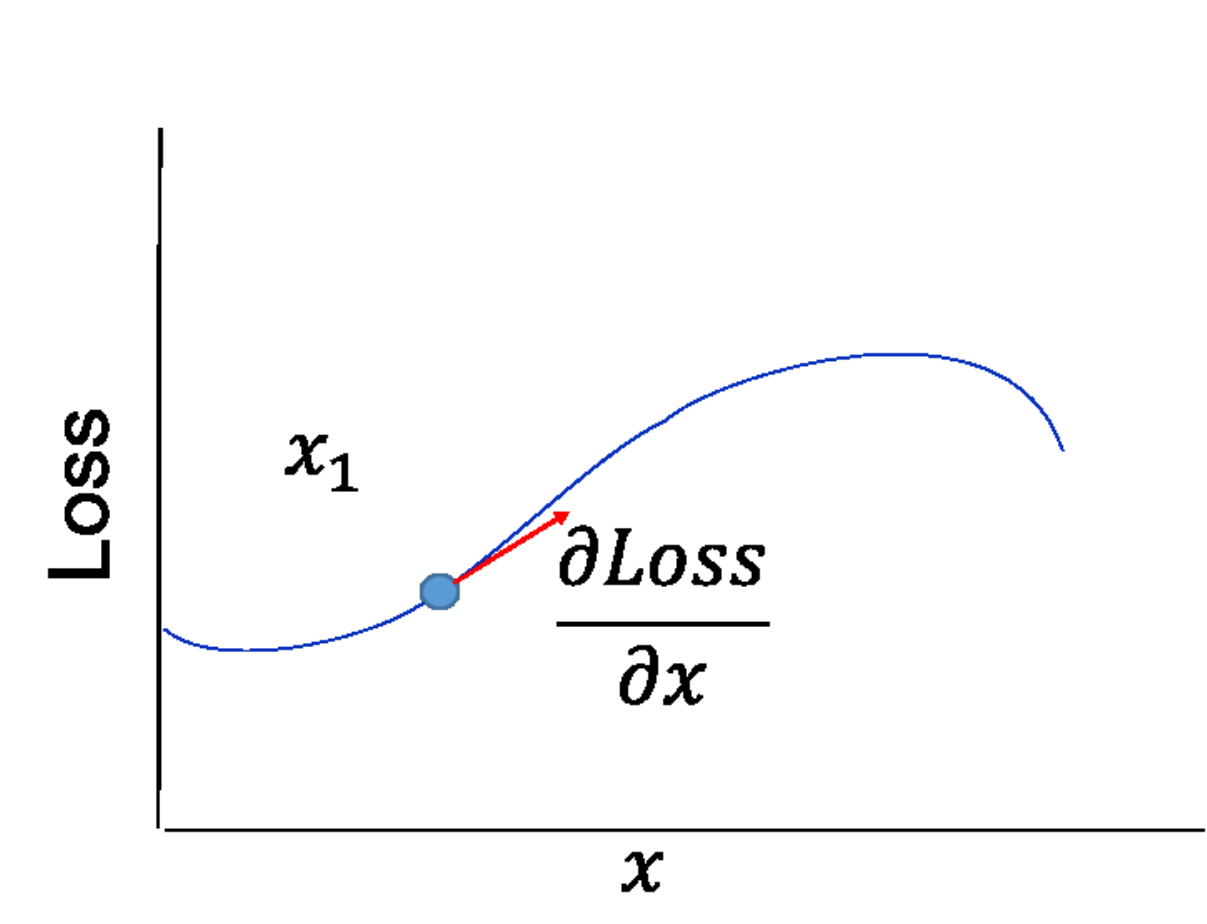} & \hspace{-2mm} \includegraphics [width=1.8in]{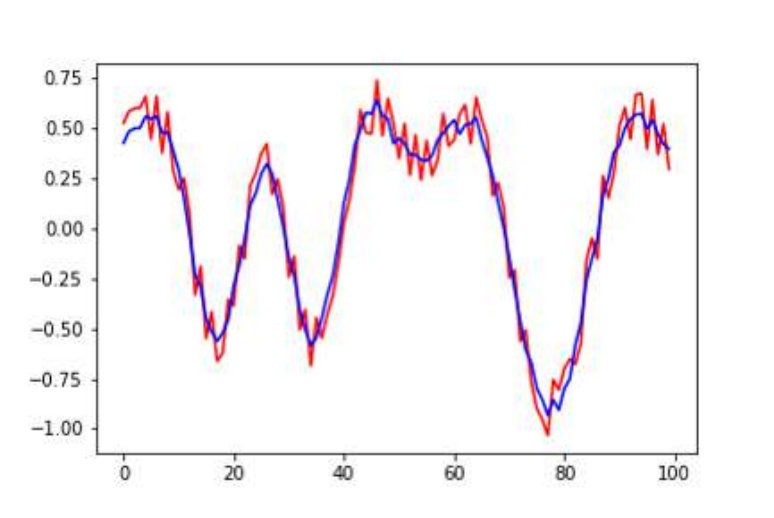}\vspace{-2mm}\\ 
{\small \textbf{A:} Basics of AdEx design}   & {\small \textbf{B:} Effect on a BPSK signal} \vspace{-2mm}
\end{tabular}
\caption{{FGSM Attack (A) and its effect on a modulated RF signal (B) }}\vspace{-4mm}   \label{fig:f0}
\end{center}
\end{figure}

As deep learning (DL) penetrates non-traditional applications, a new research direction is emerging in the field of wireless signals, aiming to emulate and possibly replace certain signal processing algorithms by DL models. Spectrum sensing, especially in the context of cognitive radio, defines most of the radio signal detection problems that are being addressed by DL. 
The approach to DL in RF domain differs from  common DL applications, such as image recognition or natural language processing, and requires knowledge of RF signal processing and wireless communications and/or radar, depending on the signal utilization. While research on adversarial examples in machine learning for images has been prolific, similar attacks on deep learning of radio frequency (RF) signals and the mitigation strategies are not addressed in the published work almost at all, with only one recent publication on RF \cite{Larsson}. Adversarial examples (AdExs) are slightly perturbed inputs that are classified incorrectly by the Machine Learning (ML) model \cite{AdExSem1}. This perturbation is achieved by mathematical processing of the signal, e.g., by adding an incremental value in the direction of the classifier’s gradient with respect to the inputs (Fig.~\ref{fig:f0}~A), or by solving a constrained optimization problem. Popular deep learning (DL) models are even more vulnerable to AdExs as DL networks learn input-output mappings that are fairly discontinuous. Consider the stop-sign images in Figure~\ref{fig:f00} \cite{GoodfellowMcDanielPapernot18}. To the human eye, they appear to be the same -- two images of a stop sign. The image on the left is an ordinary image of stop sign, while the one on the right is derived from it by adding a precise perturbation that elicits the deep classifier to classify it as a yield sign. For further details about AdExs, please see seminal work, such as \cite{AdExSem1, AdExSem2}. RF adversarial examples can cause drastic, targeted misclassification results mostly in spectrum sensing/ survey applications (e.g. BPSK mistaken for OFDM) with minimal waveform perturbation.  It is not clear if the RF AdExs maintain their effects in the physical world, i.e., when AdExs are delivered over-the-air (OTA). Our research on deep learning AdExs and proposed defense mechanisms are RF-centric, and incorporate physical-world, OTA effects.  In this work we present defense mechanisms based on statistical tests. One test to detect AdExs utilizes Peak-to-Average-Power-Ratio (PAPR) of the DL data points delivered OTA, while another statistical test uses the Softmax outputs of the DL classfier, which corresponds to the probabilities the classifier assigns to each of the trained classes. The former test leverages the RF nature of the data, and the latter is universally applicable to AdExs regardless of their origin. Both solutions are shown as viable mitigation methods to subvert adversarial attacks against communications and radar sensing systems. 
\subsubsection{Existing Work}
The research in the area of RF-based DL of the PHY layer is still embryonic \cite{PHYDeep}. Modulation recognition \emph{(ModRec)} is the most popular application of DL here. Most of the existing work is based on convolutional (CNN) architectures \cite{ConvDeepRF}.  Paper \cite{OTADeepRF} features an in-depth study on the performance of DL ModRec methods on OTA captured RF communication signals synthetically designed in Software Defined Radio (SDR). The paper \cite{OTADeepRF} demonstrates that in the ModRec context DL provides significant performance benefits compared to conventional feature extraction methods. Apart from exploring optimal DL architectures and comparing their classification accuracy with state-of-the-art performance based on signal cumulants or their cyclo-stationary properties \cite{cyclostat}, this paper contributed a publicly available dataset  \cite{DeepSigDataset}, which we exploit here to demonstrate detection and mitigation of the attacks on DL that leverage adversarial examples of RF data points.
Most current attacks are based on the gradient of a neural network’s (NN) loss function: White-box attacks use the target NN to compute the gradient; Black-box attacks use a surrogate network to approximate the gradient. 
We will be using Fast Gradient Sign Method (FGSM) \cite{AdExSem2} to illustrate our ideas. This attack takes the sign of the gradient and moves the data point $x$ one step in that direction.
\begin{align}
\nonumber &\hat{x} = x+\epsilon sign(\nabla_x J(x,y_{adv},\theta)),
\end{align}
where $x$ is the legitimate data point, and $\hat{x}$ is its adversarial example. $J(x,y_{adv},\theta)$ is the loss function of the NN (with trained parameters $\theta$) at the input $x$ evaluated for the targeted label. We denoted the targeted (adversarial classification) label as $y_{adv},$ which can be constant, or a random value, and the hyperparameter $\epsilon$ is usually a small number to limit the perturbation.

FGSM is simple and defendable (as proven in many cases of image ML), but it is also the basic principle used in iterative methods and constrained-optimization-based methods, hence representing a good reference for evaluation of new defense approaches. We will not go into details due to space constraints. Let us just mention a couple of iterative methods based on FGSM: Basic iterative method (multiple steps of FGSM) \cite{AdExFGSMIter}, Carlini-Wagner method \cite{AdExCnW} (similar but modified objective function), Projected Gradient Descent (add noise, compute gradient, step, project back) \cite{AdExPGD}.

\section{Adversarial Examples: Problem Statement}
\begin{figure}[t] 
\begin{center}
\hspace{-5mm} \includegraphics [width=3.5in]{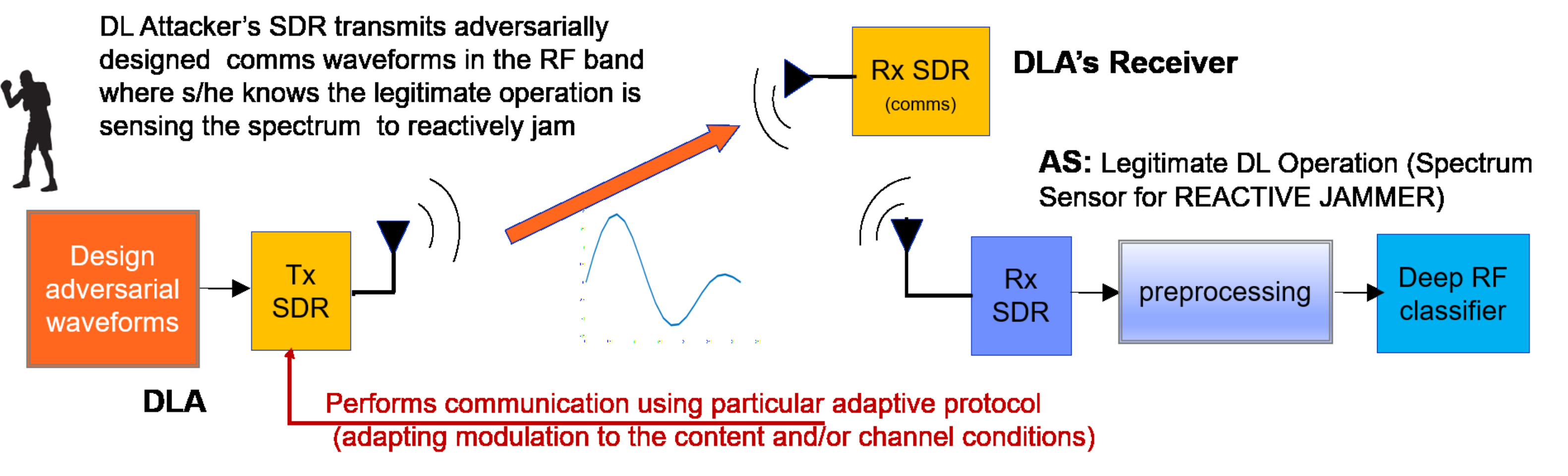} 
\caption{{Motivating scenario for an OTA attack to DL classifier via RF AdExs}}\vspace{-4mm}   \label{fig:f1}
\end{center}
\end{figure}

Consider this scenario for an OTA attack on an RF DL classifier as a motivating example (Fig.~\ref{fig:f1}). The DL attacker (DLA) is at the transmitter of a communication system. Both DLA and the attacked system (AS) are using software defined radios (SDRs), although the DLA's receiver does not have to based on an SDR. The DLA’s goal is to elicit adversarial classification $y_{adv}$ at the AS on the signal $x$ designed based on a legitimate signal of class $y$. Despite the adversarial modifications targeted to elicit classification different from $y$, the designed AdEx needs to maintain high probability of being decoded as $y$ at the attacker’s own intended receiver, hence the perturbation of $x$ is constrained. 

The AS is sensing the spectrum in order to perform reactive jamming, e.g., to jam BPSK modulated signals since many preambles are BPSK-modulated, and if the preamble gets corrupted by jamming the whole packet is lost \cite{MillerVmimo}. Moreover, packets of the control plane of most protocols (such as acknowledgments) are BPSK-modulated. Hence, the DLA would want to create the AdEx that disguises BPSK as QPSK (or other) to avoid the EW attack by the DL-based reactive jammer (i.e., AS) \cite{Wilhelm2011ReactiveJam}. Note that reactive jamming is very difficult to detect \cite{ReactXu05}, but it heavily relies on the inference based on spectrum sensing. If the inference is adversarially attacked, the jammer will be mitigated. We here take the side of the jammer in attempting to detect such adversarial attacks.

There are many methods to create AdExs. By definition, the following optimization problem describers the general approach:
\begin{align}
\nonumber &\min{||r_x||_p}\\
\nonumber s.t. &\ell(x) \neq \ell(x+r_x),\ \ and\\ 
& x+r_x \in \mathcal{X},
\end{align}
where $||\cdot||_p$ denotes the $l_p$ norm, and $\ell(x,\theta)$ is the decision rule by the NN with parameters $\theta$ evaluated at $x.$ The final constraint is somewhat arbitrary: it means that the adversarial example still belongs to the same space as the legitimate data point.

In the case of images the {\em slight perturbation} applied to a legitimate data example is expressed as visual imperceptibility by a human viewer. RF adversarial examples is a nascent field, and as such the definition of imperceptible perturbation does not exist in the literature. 
For RF signals utilized for communications we define the imperceptible perturbation as any deformation of the RF waveform that can be filtered out by a receiver, e.g. via matched filters or correction codes, such that the bit-error rate is close to that of legitimate signals. An analogous definition can be made for radar using receiver operating characteristics (ROC). 

Mitigating adversarial inputs remains an open problem, even in the image-domain.  A complicating factor in detecting AdEx attacks include variations due to the physical world. Visual adversarial perturbations and their robustness given different backgrounds, lighting, and camera resolutions is discussed in \cite{Eykholt2017RobustPA}. The diversity of RF communications, radar, and spectrum sensing systems, and complex propagation channels makes this problem in the RF domain even more complex and unique. We overcome this through proper consideration and modeling of the both hardware and the channel.

As mentioned, the defense mechanisms that we are proposing rely on statistical tests. The idea of using statistical tests to detect and discard adversarial examples is not new. Common approaches to defense against adversarial attacks (mainly evaluated on image datasets) include: Gradient masking - hiding the gradient; Preprocessing - trying to “undo” the perturbations; Detecting AdExs - looking for distribution shifts; Certified methods - proving immunity to a set of perturbations. For details please see \cite{CertifiedDef} and references therein. We focus on the uniqueness of RF data and its delivery to DL-based inference systems to defend against RF adversarial attacks. For example, the detection of AdExs using statistics designed for images does not transfer to RF, as some statistics are destroyed during OTA delivery. The effect of the channel (or interference) which is mitigated in well-designed receivers will persist at the DL classifier, thus changing the classification for both legitimate inputs and AdExs. Our first statistical test utilizes PAPR statistics, which is a widely adopted and studied metric in wireless communications, and represents a signature for a particular modulation. If the classified label of an input signal indicates a specific modulation, and its PAPR statistic indicates the opposite with high confidence, additional tests should be performed to confirm or reject suspicion. The second test that we propose is universal, and relies on the statistics of the last layer of the neural network to detect a distribution shift caused by adversarial examples. Typically, the size of the statistics affects the quality of the tests, and we compare the confidence of the tests utilizing the entire corpus of the training data versus those that utilize only the statistics of inputs whose total duration is hundreds of microseconds or less. 

Finally, adversarial training of neural networks is another common defense, and often complimentary to other methods.  It consists of generating the AdExs according to one or more attack methods, and retraining the NN with labeled  AdExs. We use adversarial training in conjunction with other mitigation and defense methods.

There are some complexities in DL of RF signals that we would like to highlight since our approach to solving those complexities impacts the presented results. Raw RF signal data is complex-valued, and traditionally split into the in-phase (I) and quadrature (Q) channels, resulting in a series of $I + jQ$ samples. Standard DL networks are not designed to handle complex-valued data, hence we must apply a transform to the real domain that preserves salient signal information. Our prior research leveraged expert feature transforms (e.g. FFTs, wavelets) to optimize performance by reducing the complexity (e.g. number of NN parameters) of the proceeding network. For this research we used interleaved I/Q samples of the DeepSig dataset comprised of synthetic data points from 24 modulation classes \cite{DeepSigDataset}. This simple transform from complex to real set, could be expressed as follows: for a data point $c$ of $k$ I/Q samples $c_1, \cdots, c_k$, where $c_{\ell} = I_{\ell} +j\ast Q_{\ell},$ the transformed vector has $2k$ real elements $\set{I_1, Q_1, I_2, Q_2, \cdots, I_k, Q_k}.$ For the Deepsig dataset that we used $k=1024,$ hence the input to the NN is a tensor of dimensions $\paren{\cdot, 2048}.$ Note that despite the conversion from complex data to the interleaved I/Q transform the accuracy of classifying four modulations represented by a subset of the DeepSig dataset gets close to 100\% if data points with $SNR \geq 14$~dB are used (see \ref{fig:f2b}). Adversarial examples with $\epsilon = 0.1$ lower the accuracy by 30-40\% on average. 
\section{Proposed Methods for Mitigation and Defense}
All the results presented here are based on a subset of the DeepSig dataset (9) – specifically, we used BPSK, QPSK, 8PSK and 16QAM (DeepSig classes 3, 4 ,5, 12). We applied the FGSM attack using the CleverHans library (12). We compared results with and without the attack using outlined defense methods, including: 1) statistical detection of AdExs by comparing the distribution of the output values at the final NN layer with and without an attack, by means of the Kolmogorov-Smirnov (KS) two-sample test (see \cite{KSref} and references therein); 2) statistical detection of AdExs by computing PAPR for each data point with and without an attack and then applying the KS test between respective PAPR statistics. 
\begin{figure}[t] 
\vspace{-1mm}
\begin{center}
\hspace{-5mm} \includegraphics [width=2.5in]{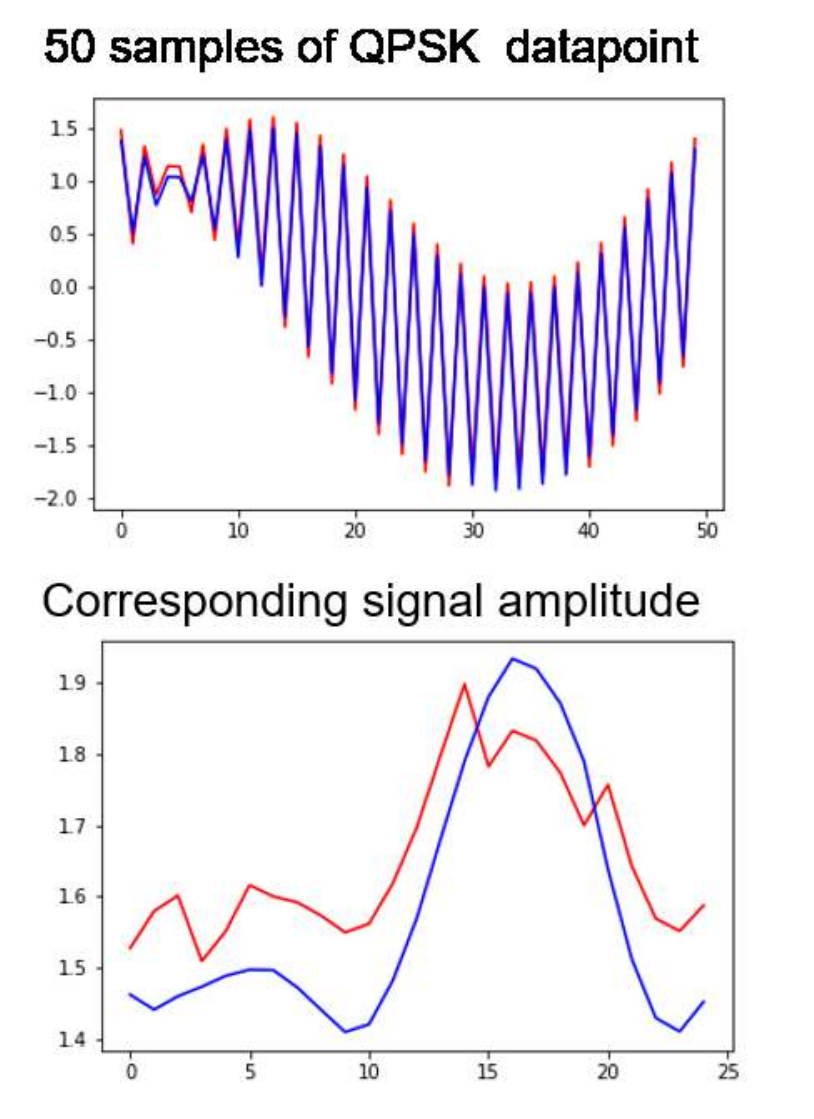} 
\caption{{Motivating example for PAPR based detection of AdExs}}\vspace{-4mm}   \label{fig:f2}
\end{center}
\end{figure}

\begin{figure}[t] 
\begin{center}
\hspace{-5mm} \includegraphics [width=3.2in]{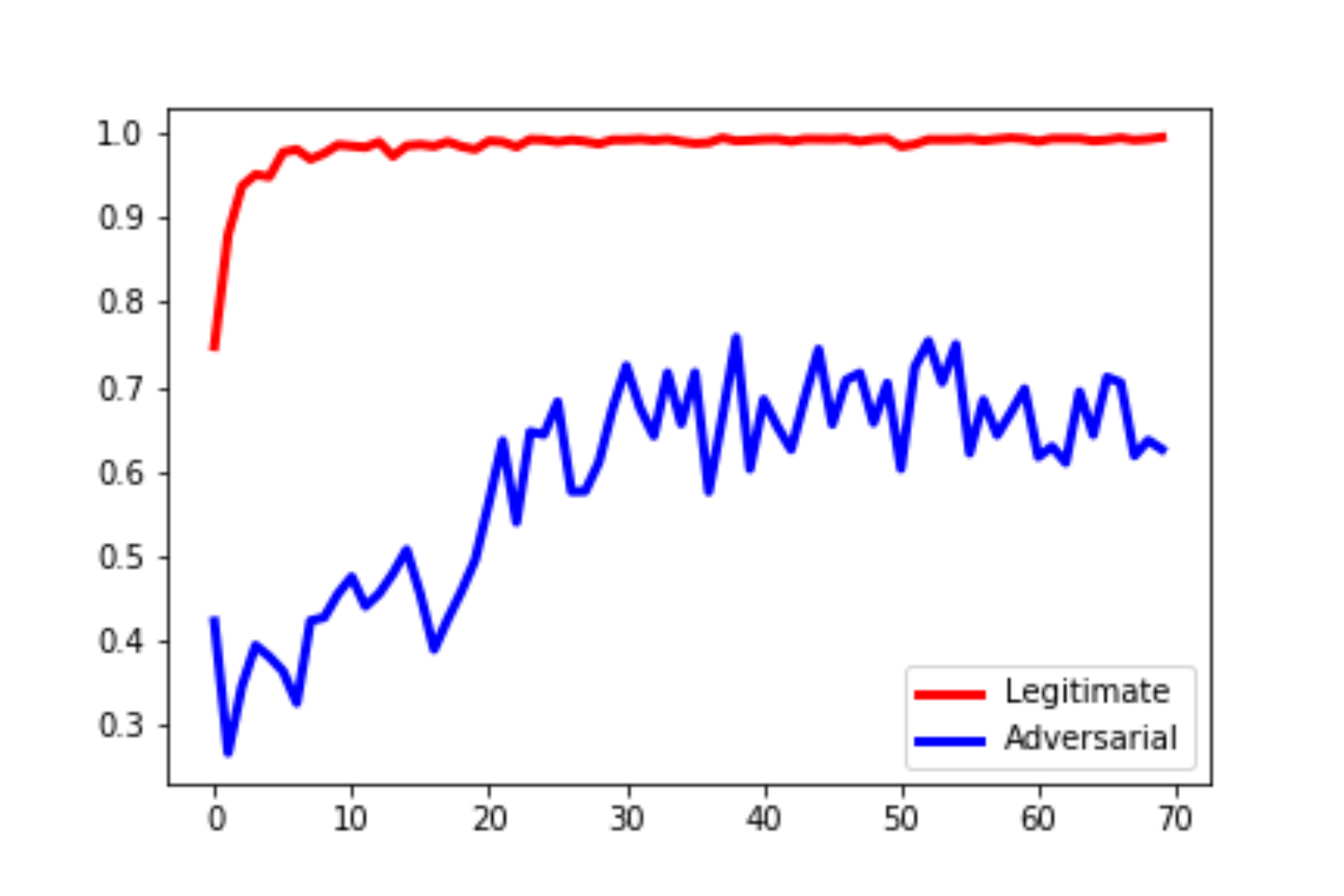} 
\caption{{Accuracy over training epochs for our CNN trained on classes 3,4,5 and 12 (BPSK, QPSK, 8-PSK,16QAM)}}\vspace{-4mm}   \label{fig:f2b}
\end{center}
\end{figure}

Let us look at the plot of the RF waveforms recovered from the legitimate data points and their adversarial examples, as presented in Figure~\ref{fig:f2} on the example of a QPSK data point. Notice that the AdExs plotted in red at the top of Figure~\ref{fig:f2} are almost visually imperceptible, yet they cause an evident drop in accuracy (see Figure~\ref{fig:f2b}). The bottom of Figure~\ref{fig:f2} shows that even though the interleaved I and Q parts (samples of a data point under attack in the top plot) are not visually changed much for a small perturbation ϵ (FGSM with ϵ= 0.1), the induced change in the amplitude of the signal (bottom) is more pronounced. Even so, using amplitude statistics to detect an adversarial attack would not be robust in an OTA scenario (due to unknown channel conditions, including the distance-based attenuation and other effects of the wireless environment). However, a metric such as Peak-to-Average-Power Ratio (PAPR) will likely be more robust for linear time-invariant channels since the channel will ‘treat equally’ both high and low amplitudes. Since PAPR depends on the data content, and 1024 samples is roughly 100 symbols in our dataset, then for BPSK this is enough to achieve statistical diversity and yield a stationary signal.  Higher order modulations may need more samples for PAPR to be a sufficiently good statistic, which the results presented below illustrate. 

\begin{figure}[t] 
\vspace{-1mm}
\begin{center}
\hspace{-5mm} \includegraphics [width=3.5in]{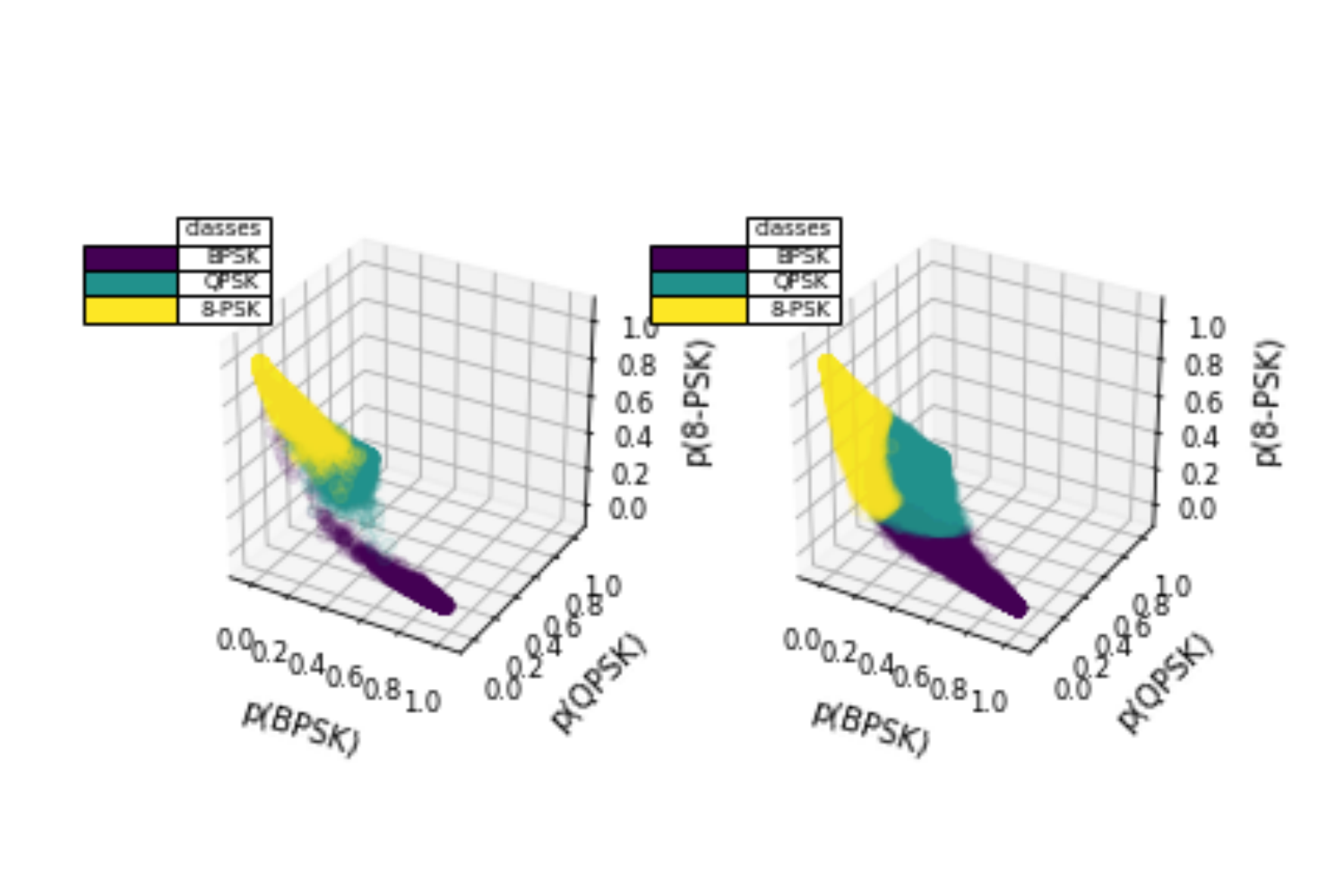}\\ 
\vspace{-8mm}\begin{tabular}{c c}
{\small \textbf{A:} Output for legitimate data}   & {\small \textbf{B:} Output for AdExs} \vspace{-2mm}
\end{tabular}
\caption{{Classification of 3 modulations (BPSK, QPSK,8-PSK) shows different distribution of output probabilities between legitimate and adversarial examples even after 2 training epochs}}\vspace{-5mm}   \label{fig:f3}
\end{center}
\end{figure}

\begin{figure}[t] 
\vspace{-1mm}
\begin{center}
\hspace{-5mm} \includegraphics [width=3.5in]{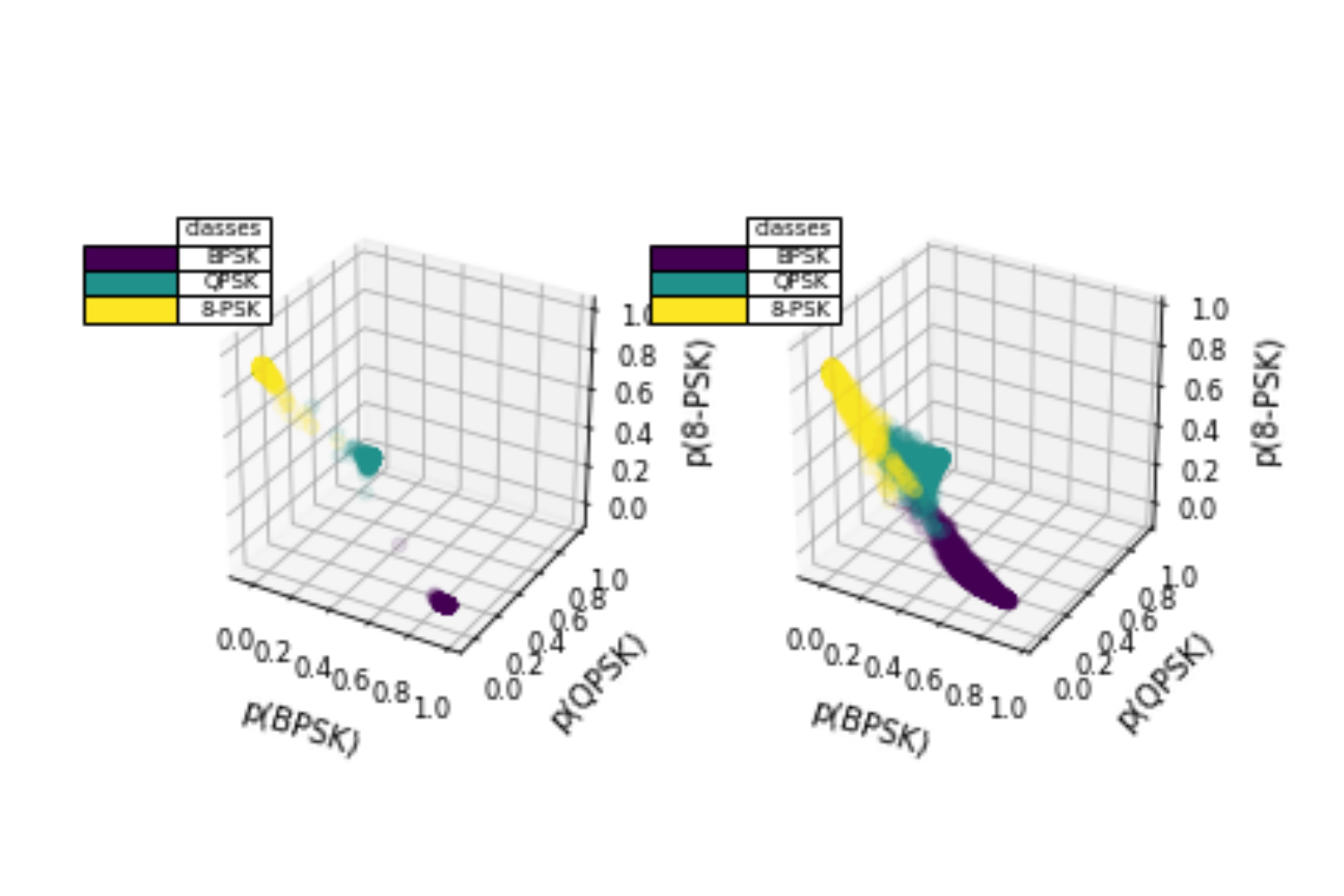}\\  
\vspace{-8mm}\begin{tabular}{c c}
{\small \textbf{A:} Output for legitimate data}   & {\small \textbf{B:} Output for AdExs} \vspace{-2mm}
\end{tabular}
\caption{{Classification of 3 modulations (BPSK, QPSK, 8-PSK) shows very different distribution of output probabilities between legitimate and adversarial examples after 20 training epochs}}\vspace{-4mm}   \label{fig:f4}
\end{center}
\end{figure}

The test that utilizes the output of the Softmax layer is motivated by Figures~\ref{fig:f3}~and~\ref{fig:f4}. For the sake of visualization, both figures are based on the 3-class classifier trained on BPSK, QPSK and 8-PSK modulation data points. Hence, the outputs of the Softmax layer are 3-dimensional vectors that are plotted in the figures for all data points utilized for training (close to 20,000, shown on the left), and for their adversarial examples, shown in the plot on the right. The elements of the vectors are values between 0 and 1, representing the probabilities of the classes. Figure~\ref{fig:f4} shows these vectors after 20 epochs of training, upon the convergence of the loss function and after the achieved accuracy exceeded 99\%. Observe that for legitimate examples, the BPSK-classified data points (purple) cluster in the area close to the vertex (1,0,0), denoting the probability of 1 for BPSK and 0 for QPSK and 8-PSK. Similarly, the points classified as QPSK and 8-PSK cluster around the vertices corresponding to probability of 1 for QPSK and 8-PSK, respectively. The left-hand side of Figure~\ref{fig:f4} shows that for the adversarial examples the vectors are distributed across a wide range of values, and the clusterization effect is lost.

Figure~\ref{fig:f3} shows the same vectors after just 2 epochs of training and no clusterization is apparent in either plot due to an untrained network, and the low accuracy for both legitimate and adversarial examples. This emphasizes the need to evaluate the adversarial effects on a completely trained network.
 
\subsection{Kolmogorov-Smirnov test for Adversarial PAPR} 
We compute PAPR for each data point and the Adversarial PAPR (APAPR) for its AdEx, and conduct statistical tests whether  PAPRs and APAPRs belong to the same probability distribution. This is a model-agnostic defense that leverages RF signal properties.  We ran 2 test experiments, with independent attack instances. Both included comparing the entire set of PAPRs with matching APAPRs, and then the comparison of small sets --- first with 50 PAPRs in the sample, and second with 200. Small sets are compared to capture the effect of the sample size on the KS test confidence (Fig.~\ref{fig:f5}). 
Columns of the tables in Fig.~\ref{fig:f5} show for each experiment  the 3 instances of the 2-sample KS test between the PAPRs and APAPRs: 1) Entire PAPR dataset per class  vs. entire APAPR dataset per predicted class; 2) A random set of 50 (or 200) PAPR values of the same class vs a random set of 50 (or 200) same class APAPR values 3)  Control instance (PAPR to PAPR, per class) with a random set of 50 (200) PAPR values vs another random set of 50 (200) legitimate PAPR values. The 3rd instance is the control experiment.
\begin{figure}[t] 
\begin{center}
\hspace{+5mm} \includegraphics [width=3.5in]{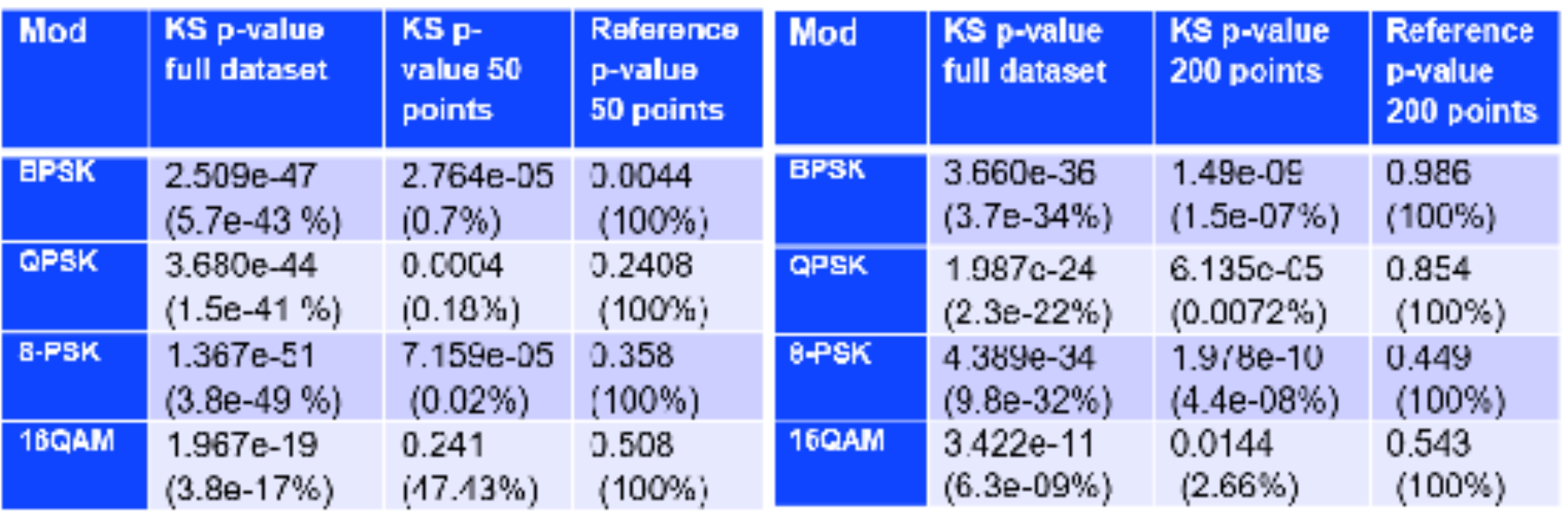}\\ 
\vspace{-2mm}\begin{tabular}{c c}
\hspace{+2mm} Table 1 & \qquad \qquad \qquad \qquad Table 2
\end{tabular}
\caption{{KS test results, based on PAPR statistics, for an experiment with a control set of 50 PAPR samples (left), and of 200 samples (right)}}\vspace{-4mm}   \label{fig:f5}
\end{center}
\end{figure}

The KS test declares the confidence (p-value) that the two sets of statistics are from the same distribution. Tables in \ref{fig:f5} display those confidence values for the 3 instances. Small confidence in the 1st and 2nd column show that AdExs are not drawn from the same distribution as the original data, and can thus be detected using this test. The 3rd column quantifies the confidence in such a claim, with one being the highest confidence. The smaller number of malicious points included in the tested sample (50) resulted in much lower detection confidence, especially for low-order modulations (as evident in Table 1 vs Table 2, 3rd column). Table 2 shows only slight improvement in the 3rd column for 16QAM as the highest order modulation needs more than 1024 samples to create good PAPR statistics. 

\subsection{Kolmogorov-Smirnov Test for Output Layer Probabilities}
This KS test is performed in the same manner as the previous one, with two experiments and three instances. The only difference is the statistics applied to evaluate the equality of probability distributions between the two samples of statistics. The statistic in matter is the empirical entropy $H$ of each data point $x$, calculated as:
\begin{align}
\nonumber &H(x) = \sum_{i=1}^{C}{p_i(x)\log{p_i(x)}},
\end{align}
where $C$ is the number of classes, and $\set{p_1(x)\cdots, p_C(x)}$ is the vector of the class probabilities at the output layer, evaluated for data point $x.$ We define the same statistic for the adversarial examples.

Figure~\ref{fig:f5} shows the two experiments, with 50 $H(x)$ control samples on the left, and with 200 $H(x)$ control samples on the left. Note that the confidence that legitimate samples and AdEx samples are from the same distribution is about zero in the first column for all modulations, and similarly for smaller sets of 50 or 200 statistics (second column). The third column shows what is the confidence when we use the legitimate data points to calculate the statistics of both KS samples. Ideally, it should be close to one, and more so for 200 than for 50 entropy values. The closer the third column is to one the more valid the confidence values in the first two columns are. We see that the increase  of the control confidence with the larger set size (in the second table) is largest for the highest-order modulation here (the last row, marked with the arrow). However, despite the values in the third column not converging to one for the presented sizes, we can better see the validity of the test if we equate those values with maximum probability of the same-distribution hypothesis (100 \%); in that case the probability that adversarial and legitimate samples from the other two columns are drawn from the same probability is close to 0\%.
 
The KS test based on the entropy of output layer probabilities will likely show different results when evaluated at a classifier that is trained on clean RF samples but receives the data points OTA, causing a distribution shift due to the channel effects. We do not evaluate these effects here, due to the lack of space. Our intention in this paper is to show that for data points representing differently modulated RF signals this test promises fairly reliable detection of the distribution shift caused by the intentional and malicious perturbations implemented by adversarial examples.

The DL NN used with both this and the PAPR test is a CNN of similar architecture as used for the 1st defense method. 
Normally trained CNN exhibits test accuracy of 0.9988 on legitimate examples and test accuracy 0.4815 on the 0.1 FGSM AdExs. The adversarially-trained CNN exhibits test accuracy of 0.9991 on legitimate examples and test accuracy 0.7659 on the 0.1-FGSM AdExs.
\begin{figure}[t] 
\begin{center}
\hspace{+2mm} \includegraphics [width=3.5in]{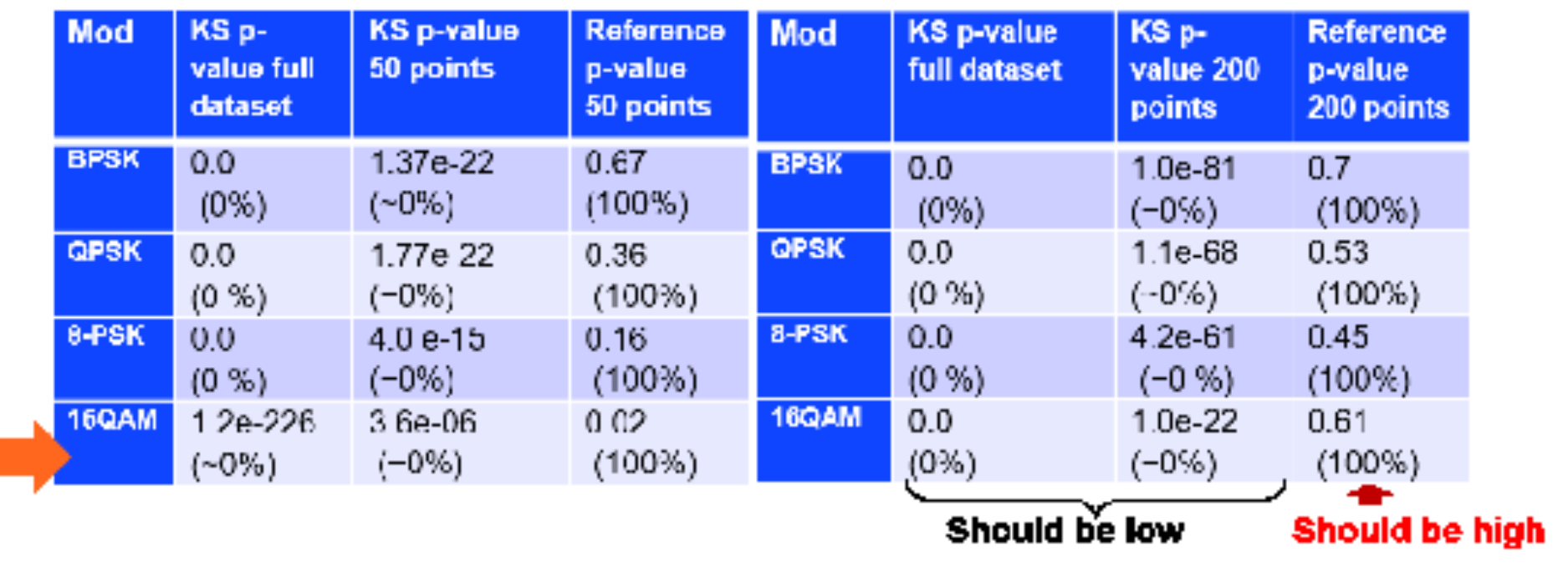}\\ 
\vspace{-2mm}\begin{tabular}{c c}
\hspace{+2mm} Table 1 & \qquad \qquad \qquad \qquad Table 2
\end{tabular}
\caption{{KS test results, based on output layer probabilities,  for an experiment with a control set of 50 output layer samples (left), and of 200 samples (right)}}\vspace{-4mm}   \label{fig:f5}
\end{center}
\end{figure}

\section{Conclusion}
We showed  that statistical tests based on PAPR values of data points, and on the probability of classes derived from the last NN layer, can detect a distribution shift created by adversarial examples. While the PAPR statistics are more robust to the over-the-air delivery of RF data points, this paper does not assert that this is true for the test based on class probabilities. Our future research will investigate the robustness of the latter.  Other statistics could be used to detect the distribution shift, which is outside the scope of this paper. We used the two methods to represent two classes of test statistics, those that are robust to wireless corruption, and those that are not. There are additional defense methods that are not based on statistical tests, whose validity should be verified in terms of robustness to corruption incurred at the receiver due to OTA delivery. As an example, we would like to mention additional methods that we pursued in this line of research. Defense against AdExs by pretraining the network by an autoencoder (AE) mitigates the deceiving effect of AdExs, but expanding the AE-based defense to a denoising AE is likely to increase its robustness against the receiver noise. The robustness of the attacks to deep learning via adversarial examples, and of the defense mechanisms against them is an open problem, and this paper is an attempt to address its manifestations and specificity in the RF domain.  

\bibliographystyle{IEEEtran}%
\bibliography{adversarial}
\end{document}